\def\year{2021}\relax
\documentclass[letterpaper]{article} 
\usepackage{aaai21}  
\usepackage{times}  
\usepackage{helvet} 
\usepackage{courier}  
\usepackage[hyphens]{url}  
\usepackage{graphicx} 

\usepackage{comment}
\usepackage{multirow}
\usepackage{bm}

\usepackage{amsmath,amssymb}

\usepackage{natbib}  
\usepackage{caption} 
\title{Weakly-Supervised Hierarchical Models for Predicting Persuasion Strategies}
\author{
    Jiaao Chen, Diyi Yang \\
}
\affiliations{
School of Interactive Computing \\
    Georgia Institute of Technology \\
    \{jchen896,dyang888\}@gatech.edu
   
}

\begin{document}

\maketitle

\begin{abstract}
Modeling persuasive language has the potential to better facilitate our decision-making processes. Despite its importance, computational modeling of persuasion is still in its infancy, largely due to the lack of benchmark datasets that can provide quantitative labels of persuasive strategies to expedite this line of research. 
To this end, we introduce a large-scale multi-domain text corpus for modeling persuasive strategies in good-faith text requests.
Moreover, we design a hierarchical weakly-supervised latent variable model that can leverage partially labeled data to predict such associated persuasive strategies for each sentence, where the supervision comes from both the overall document-level labels
and very limited sentence-level labels. Experimental results showed that our proposed method outperformed existing semi-supervised baselines significantly. We have publicly released our code at \url{https://github.com/GT-SALT/Persuasion_Strategy_WVAE}. 
\end{abstract}

\section{Introduction}
Persuasive communication has the potential to bring significant positive and pro-social factors to our society \cite{hovland1971communication}. For instance, persuasion could largely help fundraising for charities and philanthropic organizations or convincing substance-abusing family members to seek professional help. Given the nature of persuasion, it is of great importance to study how and why persuasion works in language.
Modeling persuasive language is challenging in the field of natural language understanding since it is difficult to quantify the persuasiveness of requests and even harder to generalize persuasive strategies learned from one domain to another.
Although researchers from social psychology have offered useful advice on how to understand persuasion, most of them have been conducted from a qualitative perspective \citep{bartels2006priming,popkin1994reasoning}.  Computational modeling of persuasion is still in its infancy, largely due to the lack of benchmarks that can provide unified, representative corpus to facilitate this line of research, with a few exceptions like \citep{luu2019measuring,atkinson2019gets,wang2019persuasion}.

Most existing datasets concerning persuasive text are either (1) too small (e.g., in the order of hundreds) for current machine learning models \cite{yang2019let} or (2) not representative for understanding persuasive strategies by only looking at one specific domain \cite{wang2019persuasion}. To make persuasion research and technology maximally useful, both for practical use and scientific study,
a generic and representative corpus is a must, which can represent persuasive language in a way that is not exclusively tailored to any one specific dataset or platform. To fill these gaps, building on theoretical work on persuasion and these prior empirical studies, we first introduce \textbf{a set of generic persuasive strategies and a multi-domain corpus} to understand different persuasion strategies that people use in their requests for different types of persuasion goals in various domains. 

However, constructing a large-scale dataset that contains persuasive strategies labels is often 
time-consuming and expensive.  To mitigate the cost of labeling fine-grained sentence persuasive strategy,
we then introduce \textbf{a simple but effective weakly-supervised hierarchical latent variable model} that  
leverages mainly global or document-level labels (e.g., overall persuasiveness of the textual requests)
alongside with limited sentence annotations to predict sentence-level persuasion strategies.
Our work is inspired by prior work \cite{oquab2015object} in computer vision that used the global image-level 
labels to classify local objects. Intuitively, our model is hierarchically semi-supervised, with sentence-level latent variables to reconstruct the input sentence and all latent variables of sentences are aggregated to predict document-level persuasiveness. Specifically, at the sentence-level, we utilize two latent variables representing persuasion strategies and context separately, in order to disentangle information pertaining to label-oriented and content-specific properties to do reconstructions;  at the document level, we encode those two latent variables together to predict the overall document labels in the hope that it could supervise the learning of sentence-level persuasive strategies.  
To sum up, our contributions include: 
\begin{enumerate}
    \item A set of generic persuasive strategies  based on theoretical and empirical studies and introducing a relatively large-scale dataset that includes annotations of persuasive strategies for three domains.
    \item A hierarchical weakly-supervised latent variable model to predict persuasive strategies with partially labeled data.
    \item Extensive experimental results that test the effectiveness of our models and visualize the importance of our proposed persuasion strategies.
\end{enumerate}
\section{Related Work} 
There has been much attention paid to computational persuasive language understanding \cite{guo2020opinion,atkinson2019gets,lukin2017argument,yang2017persuading,shaikh2020examining}. For instance,  \citet{Tan:2016:WAI:2872427.2883081} looked at how the interaction dynamics such as the language interplay between opinion holders and other participants predict the persuasiveness via ChangeMyView subreddit.
\citet{Althoff+al:14a} studied donations in Random Acts of Pizza on Reddit, using the social relations between recipient and donor plus linguistic factors like narratives to predict the success of these altruistic requests.
Although prior work offered predictive and insightful models, most research determined their persuasion labels or variables without reference to a taxonomy of persuasion techniques.
\citet{yang2019let} identified the persuasive strategies employed in each sentence among textual requests from crowdfunding websites in a semi-supervised manner. \citet{wang2019persuasion} looked at utterance in persuasive dialogues and annotated a corpus with different persuasion strategies such as self-modeling, foot-in-the-door, credibility, etc., together with classifiers to predict such strategies at a sentence-level. These work mainly focused on a small subset of persuasion strategies and the identification of such strategies in a specific context. Inspired by those work, we propose
a generic and representative set of persuasion strategies to capture various persuasion strategies that people use in their requests.

\newcommand{\tabincell}[2]{\begin{tabular}{@{}#1@{}}#2\end{tabular}}
\begin{table*}[ht]
\centering
\small
\begin{tabular}{r|l|l}
\hline
  \textbf{Strategy} & \textbf{Definition and Examples} & \textbf{Connection with Prior Work} \\
\hline\hline
\tabincell{l}{Commitment} &\tabincell{l}{The persuaders indicating their intentions to take acts or  justify their earlier \\decisions to convince others that they have made the correct choice. \\ e.g., \textit{I just lent to Auntie Fine's Donut Shop.} (\textit{Kiva})
} &\tabincell{l}{\emph{Commitment} (Yang et al. \citeyear{yang2019let}), \\ \emph{Self-modeling} (Wang et al. \citeyear{wang2019persuasion}), \\\emph{Commitment} (Vargheese et al. \citeyear{10.1007/978-3-030-45712-9_2})
}  \\ \hline
\tabincell{c}{Emotion} &\tabincell{l}{Making request full of emotional valence and arousal affect to influence others. \\ e.g., \textit{Guys I'm desperate.} (\textit{Borrow}) \\ \textit{I've been in the lowest depressive state of my life.} (\textit{RAOP})  } &\tabincell{l}{\emph{Ethos} \cite{carlile-etal-2018-give},\\ \emph{Emotion appeal} \cite{carlile-etal-2018-give},\\ \emph{Sentiment} (Durmus et al. \citeyear{durmus-cardie-2018-exploring}), \\\emph{Emotion words} (Luu et al, \citeyear{TACL1639}), \\ \emph{Emotion} \cite{asai-etal-2020-emotional} } \\ \hline

\tabincell{c}{Politeness} &\tabincell{l}{The usage of polite language in requests. \\
e.g., \textit{Your help is deeply appreciated!} (\textit{Borrow})} &\tabincell{l}{\emph{Politeness} (Durmus et al. \citeyear{durmus-cardie-2018-exploring}), \\ \emph{Politeness} (Althoff et al. \citeyear{Althoff+al:14a}), \\ \emph{Politeness} (Nashruddin et al. \citeyear{nashruddin2020moral})}\\ \hline
\tabincell{l}{Reciprocity} &\tabincell{l}{Responding to a positive action with another 
positive action. People are more \\likely to help if they have received help themselves. \\
e.g., \textit{I will pay 5\% interest no later than May 1, 2016.} (\textit{Borrow}) \\
\textit{I'll pay it forward with my first check.} (\textit{RAOP}) \\

} &\tabincell{l}{\emph{Reciprocity} (Althoff et al. \citeyear{Althoff+al:14a}), \\ \emph{Reciprocity} \cite{ROETHKE2020113268}, \\\emph{Reciprocity} (Vargheese et al. \citeyear{10.1007/978-3-030-45712-9_2})} \\ \hline

\tabincell{c}{Scarcity}  &\tabincell{l}{People emphasizing on the urgency, rare of their needs.\\ e.g., \textit{Need this loan urgently.} (\textit{Borrow}) \\ \textit{I haven't ate a meal in two days.} (\textit{RAOP}) \\ \textit{Loan expiring today and still needs \$650.} (\textit{Kiva})}  &\tabincell{l}{\emph{Scarcity}  (Vargheese et al. \citeyear{10.1007/978-3-030-45712-9_2}), \\ \emph{Scarcity} \cite{ yang2019let}, \\  \emph{Scarcity} \cite{lawson2020email}}   \\\hline\hline
\tabincell{l}{Credibility} &\tabincell{l}{The uses of credentials impacts to establish
credibility and earn others’ trust. \\
e.g., \textit{Can provide any documentation needed.} (\textit{Borrow})  \\
\textit{She has already repaid 2 previous loans.} (\textit{Kiva})
} &\tabincell{l}{ \emph{Credibility appeal} \cite{wang2019persuasion}, \\ \emph{Social Proof} \cite{ROETHKE2020113268}, \\ \emph{Social Proof} (Vargheese et al. \citeyear{vargheese2020exploring})} \\ \hline
\tabincell{l}{Evidence} &\tabincell{l}{Providing concrete facts or evidence for
the narrative or request. \\
e.g. \textit{My insurance was canceled today.} (\textit{Borrow}) \\
\textit{There is a Pizza Hut and a Dominos near me.} (\textit{RAOP}) \\
\textit{\$225 to go and 1 A+ member on the loan.} (\textit{Kiva})} &\tabincell{l}{\emph{Evidentiality } (Althoff et al. \citeyear{Althoff+al:14a}),\\ \emph{Evidence} \cite{carlile-etal-2018-give}, \\ \emph{Evidence} \cite{stab2014identifying},  \\ \emph{Concreteness}  \cite{ yang2019let} \\ \emph{Evidence}   (Durmus et al. \citeyear{durmus-cardie-2018-exploring})} \\\hline 
\tabincell{c}{Impact} &\tabincell{l}{Emphasizing the importance or impact of the request. \\ e.g., \textit{I will use this loan to pay my rent.} (\textit{Borrow}) \\ \textit{This loan will help him with his business.} (\textit{Kiva})} &\tabincell{l}{\emph{Logos} \cite{carlile-etal-2018-give}, \\ \emph{Logic appeal} \cite{wang2019persuasion} \\ \emph{Impact} \cite{yang2019let}} \\ \hline

\end{tabular} \caption{The generic taxonomy of persuasive strategies, their definitions, example sentences, and connections with prior work.}\label{Tab:def}
\end{table*}


Recently many semi-supervised learning 
approaches have been developed for natural language processing, including  adversarial 
training~\cite{miyato2016adversarial}, variational 
auto-encoders~\cite{kingma2014semi,yang2017improved,gururangan2019variational}, consistency training~\cite{xie2020unsupervised,Chen2020SemisupervisedMV,chen-etal-2020-mixtext} and 
various pre-training techniques~\cite{kiros2015skip,dai2015semi}. The contextual 
word representations~\cite{peters2018deep, devlin-etal-2019-bert} have emerged as powerful 
mechanisms to make use of large scale unlabeled data. Most of these prior works focus on semi-supervised learning, in which the labels are partially available and the supervisions for labeled and unlabeled data are both on the sentence-levels. In contrast, our
work is hierarchical weakly supervised and we aim to predict \textbf{sentence-levels labels, not document-level persuasiveness}.
To our best knowledge, weakly supervised learning has  been explored much less in
natural language processing except for a few recent 
work~\cite{lee2019latent,min2019discrete} in question answering.
There are a few exceptions: \citet{yang2019let}
 utilized a small amount of hand-labeled sentences together with a large number of requests automatically labeled at the document level for text classification. 
 \citet{pryzant2017predicting} proposed an adversarial objective to learn text features highly predictive of advertisement outcomes.
Our work has an analog task in computer vision--weakly supervised
image segmentation~\cite{papandreou2015weakly,pinheiro2015image}-- which uses image labels or bounding boxes information to
predict pixel-level labels. Similar to image segmentation, obtaining global/document/image level labels for persuasive understanding is
much cheaper than local/sentence/pixel level labels.
Different from multi-task learning where models have full supervisions in each task, our proposed model is fully supervised at the document level while partially supervised at the sentence level.
\section{Persuasion Taxonomy and Corpus}
Previous work modeling persuasion in language either focus on a small subset of strategies or look at a specific platform, hard to be adapted to other contexts.
To fill this gap, we propose a set of generic persuasive strategies based on 
widely used persuasion models from social psychology.
Specifically, we leverage Petty and Cacioppo's elaboration likelihood model (\citeyear{petty1986elaboration}) and Chiaken's social information processing model \citep{chaiken1980heuristic}, 
which suggest that people process information in two ways: either performing a relatively deep analysis of the quality of an argument or relying on some simple superficial cues to make decisions \citep{cialdini20016}. 
Guided by these psychology insights, we examine the aforementioned computational studies on persuasion and argumentation \cite{wang2019persuasion, yang2019let, durmus-cardie-2018-exploring, vargheese2020exploring, carlile-etal-2018-give}, and further synthesize these theoretical and practical tactics into eight unified categories:
\emph{Commitment, Emotion, Politeness, Reciprocity, Scarcity} that allow people to use simple inferential rules to make decisions, 
and 
\emph{Credibility, Evidence, Impact} that require people to evaluate the information based on its merits, logic, and importance. 
As shown in Table \ref{Tab:def}, 
our taxonomy \textbf{distilled, extended}, and \textbf{unified} existing persuasion strategies. 
Different from prior work that introduced domain-specific persuasion tactics with limited generalizability, our generic taxonomy can be easily plugged into different text domains, making large-scale understanding of persuasion in language across multiple contexts comparable and replicable. 

\subsection{Dataset Collection \& Statistics}
We collected our data from three different domains related to persuasion. (1) \textbf{Kiva}\footnote{\url{www.kiva.org}} is a peer-to-peer philanthropic lending platform where persuading others to make loans is a key to success (no interest),
(2) subreddit  ``r/Random$\_$Acts$\_$of$\_$Pizza\footnote{\url{www.reddit.com/r/Random_Acts_Of_Pizza}}'' (\textbf{RAOP}) where members write requests to ask for free pizzas (social purpose, no direct money transaction), and
(3) subreddit ``r/borrow\footnote{\url{www.reddit.com/r/borrow}}'' (\textbf{Borrow}) that focuses on writing posts to borrow money from others (with interest). After removing personal and sensitive information, we obtained 40,466 posts from Kiva, 18,026 posts from RAOP, and 49,855 posts from Borrow. 

We sampled 5\% documents with document length ranging from 1 to 6 from Kiva, 1 to 8 from RAOP and 1 to 7 from Borrow to annotate, as documents with at most 6 sentences account for 89\% in Kiva, 86\% posts in RAOP has no more than 8 sentences, and 85\% posts in Borrow has at most 7 sentences. We recruited four research assistants to label persuasion strategies for each sentence in sampled documents. Definitions and examples of different persuasion strategies were provided, together with a training session where we asked annotators to annotate a number of example sentences and walked them through any disagreed annotations.  To assess the reliability of the annotated labels, we then asked them to annotate the same 100 documents with 400 sentences and computed Cohen's Kappa coefficient to measure inter-rater reliability. We obtained an average score of 0.538 on Kiva, 0.613 on RAOP, and 0.623 on Borrow, which indicates moderate agreement \cite{articlecohen}. 
Annotators then annotated the rest 1200 documents by themselves independently.

\begin{figure*}[t]
\centering
\includegraphics[width=2\columnwidth]{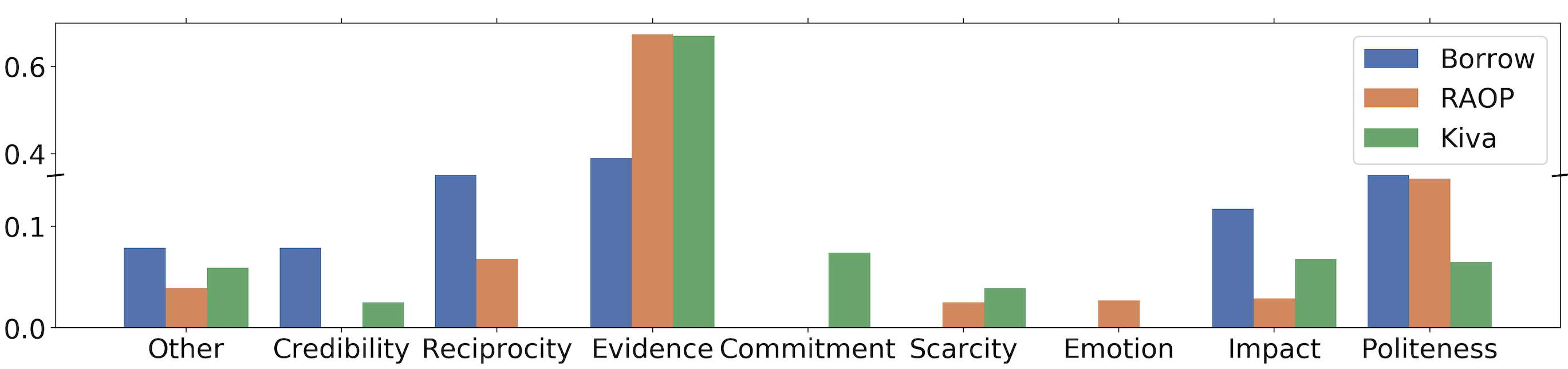}
\caption{The distribution of each persuasion strategy in three annotated three datasets.}\label{Fig:dist}
\end{figure*}

The dataset statistics 
are shown in Table \ref{Tab:statics}, and the sentence-level label distribution in each dataset is shown in Figure \ref{Fig:dist}.
We merge rare strategies into the Other category. 
Specifically, we merge Commitment, Scarcity, and Emotion in Borrow,  Credibility and Commitment in RAOP, Reciprocity and Emotion in Kiva, as Other.  We utilized whether the requester received pizzas or loans from the subreddits as the document-level labels for RAOP and Borrow. 30.1\% of people successfully got pizzas on RAOP and 48.5\% of people received loans on Borrow. In Kiva, 
we utilized the number of people who lent loans as the document-level labels. The numbers are further labeled based on buckets: $[1,2)$, $[2,3)$, $[3,4)$, $[4,\infty)$,
accounting for 44.1\%, 20.3\%, 12.4\% and 33.2\% of all documents. 
\vspace{-0.05in}

\begin{table*}[t!]
\centering
\small
\begin{tabular}{c|c|c|c|c|c}
\hline
                & \textbf{\# Docs}  & \textbf{\begin{tabular}[c]{@{}c@{}}\# Sents \\w/ label\end{tabular}} & \textbf{\begin{tabular}[c]{@{}c@{}}\# Sents \\w/o label\end{tabular}} & \textbf{Doc Labels}     & \textbf{Sent Labels}                                                                                           \\ \hline
\textbf{Borrow} & 49,855                 & 5,800                                                                & 164,293                                                                      & Success or not          & \begin{tabular}[c]{@{}c@{}}Evidence, Impact, Politeness, Reciprocity, Credibility\end{tabular}            \\ \hline
\textbf{RAOP}   &  18,026                & 3,600                                                               &    77,517                                                                   & Success or not          & \begin{tabular}[c]{@{}c@{}}Evidence, Impact, Politeness, Reciprocity, Scarcity,  Emotion\end{tabular}     \\ \hline
\textbf{Kiva}   &  40,466                & 6,300                                                               &   135,330                                                                    & \# People loaned & \begin{tabular}[c]{@{}c@{}}Evidence, Impact, Politeness, Credibility, Scarcity, Commitment\end{tabular} \\ \hline
\end{tabular}
\caption{Dataset statistics. For strategies that are rare, we merged them into an \emph{Other} category.  }\label{Tab:statics}
\end{table*}

\section{Method}
\begin{figure*}[t]
\centering
\includegraphics[width=2\columnwidth]{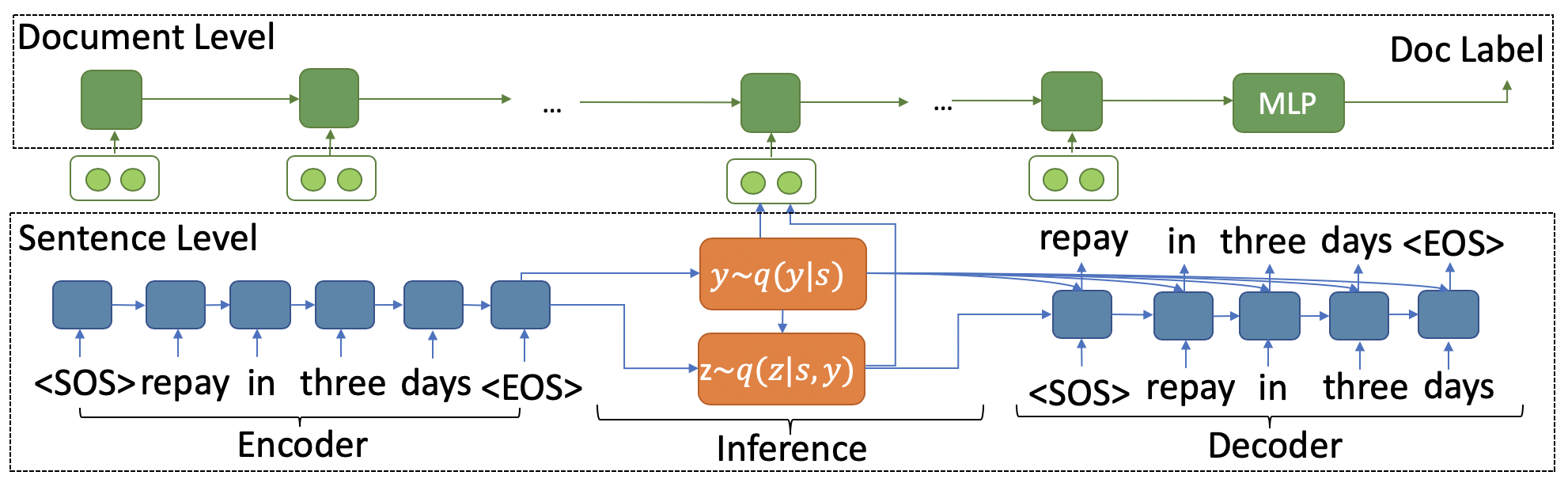}
\caption{\small{Overall architecture. At sentence-level, the input sentences are first encoded into two latent variables: $y$ representing strategies and $z$ containing context information; the decoder reconstructs the input sentences. At document-level, a predictor network aggregates the latent variables within the input document to predict document-level labels. For labeled documents, labels are directly used for the reconstruction and prediction; for unlabeled ones, latent variables $y$ are used.}}\label{Fig:model}
\end{figure*}

To alleviate the dependencies on labeled data, we propose a hierarchical weakly-supervised latent variable model to leverage partially labeled data to predict sentence-level persuasive strategies. 
Specifically, we introduce a sentence-level latent variable model to reconstruct the input sentence and predict the sentence-level persuasion labels spontaneously, supervised by the global or document-level labels (e.g., overall persuasiveness of the documents). 
The overall architecture of our method is shown in Figure \ref{Fig:model}.
\vspace{-0.05in}
\subsection{Weakly Supervised Latent Model} 
Given a corpus of $N$ documents $\mathbf{D} = \{\mathbf{d}_i\}_{i=1}^{N}$, where each document $\mathbf{d}$ consists of $M$ sentences $\mathbf{d}_i = \{\mathbf{s}_i^j \}_{j=1}^M$. For each document $\mathbf{d}_i \in \mathbf{D}$, its document level label is denoted as
$\mathbf{t}_i$,
representing the overall persuasiveness of the documents. We divide the corpus into two parts: $\mathbf{D} = \mathbf{D}_L \cup \mathbf{D}_U$, where $\mathbf{D}_L$ 
($\mathbf{D}_U$) denotes the set of documents with (without) {\em sentence} labels. For each document $\mathbf{d}_i \in \mathbf{D}_L$, the corresponding sentence labels are $\{\mathbf{y}_i^j\}_{j=1}^{M}$, where $\mathbf{y}_i^j \in \mathbf{C} = \{c_k\}_{k=1}^K$ and represents
the persuasive strategy of a given sentence.  
In many practical scenarios, getting document-level labels $\{\mathbf{t}_i\}$ is much easier and cheaper than the fine-grained sentence
labels $\{\mathbf{s}_i^j\}$ since the number of sentences $M$ in a document $\mathbf{d}_i$ can be very large. Similarly, in our setting, 
the number of 
documents with fully labeled sentences is very limited, i.e., $|\mathbf{D}_L| \ll |\mathbf{D}|$. To this end, we introduce a novel hierarchical
weakly supervised latent variable model that can leverage both the document-level labels and the small amount of sentence-level labels
to discover the sentence persuasive strategies. Our model is \textbf{weakly supervised} since we will utilize document labels to facilitate the learning of sentence persuasive strategies. The intuition is that global documents labels of persuasiveness
carry useful information of local sentence persuasive strategies, thus can provide supervision in an \textbf{indirect} manner. 

\subsubsection{Sentence Level VAE} 
Following prior work on semi-supervised variational autoencoders (VAEs) \cite{kingma2013autoencoding}, for an input sentence $\mathbf{s}$, we assume a graphical model whose latent representation contains a continuous vector $\mathbf{z}$, denoting the content of a sentence, and a discrete persuasive strategy label $\mathbf{y}$:
\begin{align}
    p(\mathbf{s},\mathbf{z},\mathbf{y}) = p(\mathbf{s}|\mathbf{z},\mathbf{y})p(\mathbf{z})p(\mathbf{y})
\end{align}

To learn the semi-supervised VAE, we optimize the variational lower bound as our learning objective. 
For unlabeled sentence, we maximize the evidence lower bound as:
\begin{equation}
    \begin{aligned}
         \log p(\mathbf{s}) 
        &\geq \mathbb{E}_{\mathbf{y} \sim q(\mathbf{y}|\mathbf{s})} [\mathbb{E}_{\mathbf{z} \sim q(\mathbf{z}|\mathbf{s}, \mathbf{y})}[\log p(\mathbf{s}|\mathbf{z}, \mathbf{y})] \\
        & \quad - \text{KL}[q(\mathbf{z}|\mathbf{s},\mathbf{y})||p(\mathbf{z})]]  - \text{KL}[q(\mathbf{y}|\mathbf{s}) || p(\mathbf{y})]
    \end{aligned}
\end{equation}
where $p(\mathbf{s}|\mathbf{y},\mathbf{z})$ is a decoder (generative network) to reconstruct input sentences 
and $q(\mathbf{y}|\mathbf{s})$ is an encoder (an inference or a predictor network) to predict sentence-level labels.

For labeled sentences, the variational lower bound is:
\begin{equation}
    \begin{aligned}
        \log  p(\mathbf{s}, \mathbf{y}) 
        &\geq \mathbb{E}_{\mathbf{z} \sim q(\mathbf{z}|\mathbf{s}, \mathbf{y})} [\log p(\mathbf{s}|\mathbf{z},\mathbf{y})] \\
        &\quad - \text{KL}[q(\mathbf{z}|\mathbf{s},\mathbf{y})||p(\mathbf{z})] + \text{constant}
    \end{aligned}
\end{equation}
In addition, for sentences with labels, we also update the inference network $q(\mathbf{y}|\mathbf{s})$ via
minimizing the cross entropy loss $\mathbb{E}_{(\mathbf{s},\mathbf{y})}[- \log q(\mathbf{y}|\mathbf{s})]$ directly.

\subsubsection{Document Level VAE}
Different from sentence-level VAEs, we model the input document $\mathbf{d}$ with sentences $\{\mathbf{s}^j\}_{j=1}^M=\mathbf{s}^{1:M}$ as a whole and assume that the document-level label $\mathbf{t}$ depends on the sentence-level latent variables.
Thus we obtain the document-level VAE model as: 
\begin{equation}
    \begin{aligned}
    p(\mathbf{d}, \mathbf{t}, \mathbf{y}, \mathbf{z}) =  p(\mathbf{d},\mathbf{t}|\mathbf{y}, \mathbf{z}) \prod_{j=1}^M p(\mathbf{y}^j) \prod_{j=1}^Mp(\mathbf{z}^j)
   \end{aligned}
\end{equation}
where $p(\mathbf{d},\mathbf{t}|\mathbf{y}^{1:M}, \mathbf{z}^{1:M})$ is the generative model for all sentences
in the document $\mathbf{d}$ and the document label $\mathbf{t}$.

For simplicity, we further assume conditional independence between the sentences $\mathbf{s}^{1:M}$ in  $\mathbf{d}$ and its label $\mathbf{t}$ given the latent variables: $p(\mathbf{d},\mathbf{t}|\mathbf{y}^{1:M}, \mathbf{z}^{1:M}) =  p(\mathbf{t}|\mathbf{y}^{1:M}, \mathbf{z}^{1:M})  \prod_{j=1}^M p(\mathbf{s}^j|\mathbf{y}^j, \mathbf{z}^j).$
\noindent
Since the possible number of the sentence label combinations is huge, simply computing the marginal probability becomes intractable. 
Thus we optimize the evidence lower bound. 
By using mean field approximation \cite{DBLP:journals/corr/abs-1802-06126}, we factorize the posterior distribution as: $q(\mathbf{z}^{1:M}, \mathbf{y}^{1:M} | \mathbf{d}, \mathbf{t})   = \prod_{j=1}^M q(\mathbf{z}^j|\mathbf{y}^j,\mathbf{s}^j, \mathbf{t}) \prod_{j=1}^M q(\mathbf{y}^j|\mathbf{s}^j, \mathbf{t})$.
That is, the posterior distribution of latent variables $\mathbf{y}^j$ and $\mathbf{z}^j$ only depends on
the sentence $\mathbf{s}^j$ and the document label $\mathbf{t}$.
For documents without sentence labels, the evidence lower bound is:
\begin{equation}
    \begin{aligned}
    & \log p(\mathbf{d}, \mathbf{t}) \geq \mathbb{E}_{\mathbf{y} \sim q(\mathbf{y}|\mathbf{s}, \mathbf{t})} [\mathbb{E}_{\mathbf{z} \sim q(\mathbf{z}|\mathbf{s}, \mathbf{y}, \mathbf{t})} [\log p(\mathbf{t}|\mathbf{y}, \mathbf{z})\\
    & \quad + \sum_{i=1}^N \log p(\mathbf{s}^j|\mathbf{y}^j, \mathbf{z}^j)] - \sum_{j=1}^M \text{KL}[q(\mathbf{z}^j|\mathbf{s}^j, \mathbf{y}^j, \mathbf{t})||p(\mathbf{z}^j)]]  \\
    & \quad - \sum_{j=1}^M \text{KL}[q(\mathbf{y}^j|\mathbf{s}^j, \mathbf{t})||p(\mathbf{y}^j)] = U(\mathbf{d}, \mathbf{t}) \\
\end{aligned}
\end{equation}

For document with sentence labels, the variational lower bound can be adapted from above as:
\begin{equation}
    \begin{aligned}
       & \log p(\mathbf{d}, \mathbf{t}, \mathbf{y}) \geq \mathbb{E}_{\mathbf{z} \sim q(\mathbf{z}|\mathbf{s}, \mathbf{y}, \mathbf{t})} [\log p(\mathbf{t}|\mathbf{y}, \mathbf{z}) \\
    &\quad  + \sum_{i=1}^N \log p(\mathbf{s}^j|\mathbf{y}^j, \mathbf{z}^j)] -\sum_{j=1}^M \text{KL}[q(\mathbf{z}^j|\mathbf{s}^j, \mathbf{y}^j, \mathbf{t})||p(\mathbf{z}^j)] \\
    &\quad = L(\mathbf{d}, \mathbf{t}, \mathbf{y}) + \text{constant}
    \end{aligned}
\end{equation}

Combining the loss for document with and without sentence labels,
we obtain the overall loss function:
\begin{equation}
    \begin{aligned}
    L =&  \quad \mathbb{E}_{\mathbf{d} \in \mathbf{D}_U} U(\mathbf{d}, \mathbf{t}) +  \mathbb{E}_{\mathbf{d} \in \mathbf{D}_L} L(\mathbf{d}, \mathbf{t}, \mathbf{y}^{1:M}) \\
    & \quad + \alpha \cdot \mathbb{E}_{\mathbf{d} \in \mathbf{D}_L} \prod_{j=1}^{M} \log q(\mathbf{y}^j| \mathbf{s}^j, \mathbf{t})
    \end{aligned}
\end{equation}
Here,  $\mathbb{E}_{\mathbf{d} \in \mathbf{D}_L} \prod_{j=1}^{M} \log q(\mathbf{y}^j| \mathbf{s}^j, \mathbf{t})$
represents the discriminative loss for sentences with labels and $\alpha$ controls the trade-off between generative loss and discriminative
loss \footnote{The influence of $\alpha$ is discussed in Section 4 in Appendix.}.

Compared to sentence-level VAE (S-VAE) that only learns sentence representation via a generative network $p(\mathbf{s}|\mathbf{y,z})$, document-level VAE utilizes the contextual relations between sentences by aggregating multiple sentences in a document and further predicting document-level labels via  a predictor network $p(\mathbf{t}|\mathbf{y}^{1:M}, \mathbf{z}^{1:M})$. 
Document-level weakly supervised VAE (WS-VAE) incorporates both direct sentence-level supervision and indirect document-level supervision to better make use of unlabeled sentences, thus can further help the persuasion strategies classification.
Note that our hierarchical weakly-supervised latent variable model presents a generic framework to utilize dependencies between sentence-level and document-level labels, and can be easily adapted to other NLP tasks where document-level supervision is rich and sentence-level supervision is scarce.

\subsection{Training Details}
In practice, we parameterize the inference network $q(\mathbf{y} | \mathbf{s}, \mathbf{t})$ and $q(\mathbf{z} |\mathbf{y}, \mathbf{s}, \mathbf{t})$ using a LSTM or a BERT which encodes the sentences (and document label) to get the posterior distribution.
We used another LSTM as the decoder to model the the generative network $p(\mathbf{s}|\mathbf{z}, \mathbf{y})$. At the document level,
each sentence's content vector and strategy vector is fed as input to a LSTM to model the predictor network $p(\mathbf{t} | \mathbf{z}^{1:M}, \mathbf{y}^{1:M})$. 



\noindent
\textbf{Reparametrization:}
It is challenging to back-propagate through random variables as it involves non-differentiable sampling procedures. For latent variable $\mathbf{z}$, we utilized the reparametrization technique proposed by \citet{kingma2013autoencoding} to re-parametrize the Gaussian random variable $\mathbf{z}$ as $\mathbf{\mu} + \mathbf{\sigma} \epsilon$, where $\epsilon \sim N(0, I)$, $\mu$ and $\sigma$ are deterministic and differentiable.
For discrete latent variable $\mathbf{y}$, we adopted Gumbel softmax \cite{45822} 
to approximate it continuously:
\begin{equation*}
    y_{k}=\frac{\exp \left(\left(\log \left(\pi_{k}\right)+g_{k}\right) / \tau\right)}{\sum_{k=1}^{K} \exp \left(\left(\log \left(\pi_{k}\right)+g_{k}\right) / \tau\right)}
\end{equation*}
where $\pi_{1:K}$ are the probabilities of a categorical distribution, $g_k$ follows Gumbel$(0, 1)$ and $\tau$ is the temperature. The approximation is accurate when $\tau \to 0$ and smooth when $\tau > 0$. We gradually decrease $\tau$ in the training process.

\noindent
\textbf{Prior Estimation:}
Classical variational models usually assume simple priors such as uniform distributions. 
We performed a Gaussian kernel density estimation over training data to estimate the prior for $\mathbf{y}$, and assumed the latent variable $z$ follows a standard Gaussian distribution.
\section{Experiment and Result}
\begin{table}[t]
\centering
\begin{tabular}{c|c|c|c}
\hline
\textbf{Dataset} &\textbf{Train} & \textbf{Dev}   & \textbf{Test}  \\ \hline 
Borrow  &900 &400 &400 \\
RAOP   &300 &200 &300 \\
Kiva  &1000 &400 &400\\
\hline 

\end{tabular}\caption{Split statistics about train, dev, and test set.}
\label{Tab:split}
\end{table}

\begin{table*}[ht]
\centering
\begin{tabular}{c|c|c|c|c|c||c}
\hline
\multicolumn{1}{l|}{\multirow{2}{*}{\textbf{Dataset}}} &\multicolumn{1}{c|}{\multirow{2}{*}{\textbf{Model}}} & \multicolumn{4}{c||}{\textbf{Sentence-level Persuasion Strategy Prediction F1 Score}}                                                                           & \multicolumn{1}{|l}{\multirow{2}{*}{\textbf{Doc-Level Accuracy}}} \\ \cline{3-6}
\multicolumn{1}{l|}{}  & {}                      & \multicolumn{1}{c|}{\textbf{20}} & \multicolumn{1}{c|}{\textbf{50}} & \multicolumn{1}{c|}{\textbf{100}} & \multicolumn{1}{c||}{\textbf{Max}} &                          \\ \hline

\multirow{6}{*}{Kiva} & {LSTM} &{$26.1 \pm 0.8$} &{$37.6 \pm 1.0$} &{$43.3 \pm 1.0$} &{$54.6 \pm 2.0$} &- \\
&{SH-Net} &{$29.1 \pm 0.4$} &{$38.8 \pm 0.9$} &{$43.4 \pm 0.9$} &{$54.8 \pm 0.9$} &{$34.8 \pm 1.0$} \\
&{BERT} &{$28.6 \pm 4.0$} &{$38.5 \pm 0.7$} &{{$44.6 \pm 3.0$}} &{{$57.0 \pm 1.0$}} &- \\
\cline{2-7} 
&{S-VAE} &{$30.9 \pm 1.0$} &{$40.3 \pm 0.7$} &{$43.6 \pm 0.9$} &{$55.7 \pm 1.0$} &-\\
&{WS-VAE} &{$31.5 \pm 0.8$} &{{$40.9 \pm 1.0$}} &{$44.0 \pm 1.0$} &{$55.4 \pm 0.8$} &{$35.5 \pm 1.0$}\\ 
&{WS-VAE-BERT} &\bm{$34.2 \pm 0.2$} &\bm{{$43.0 \pm 0.9$}} &\bm{{$45.2 \pm 0.9$}} &\bm{{$59.1 \pm 0.9$}} &\bm{$36.7 \pm 2.0$}\\ \hline \hline
\multirow{6}{*}{RAOP} & {LSTM} &{$28.5 \pm 1.0$} &{$37.7 \pm 1.0$} &{$42.5\pm 1.0$} &{$47.8 \pm 0.9$} &-\\
&{SH-Net} &{$30.0 \pm 1.0$} &{$39.1 \pm 1.0$} &{$42.8 \pm 1.0$}&{$48.1 \pm 1.0$} &{$66.6 \pm 1.0$} \\
&{BERT} &{$30.6 \pm 2.0$} &{$39.5 \pm 2.0$} &{$43.4 \pm 2.0$} &{{$54.0 \pm 1.0$}} &- \\\cline{2-7} 
&{S-VAE} &{$31.7 \pm 0.7$} &{{$40.1 \pm 1.0$}} &{$43.2 \pm 1.0$} &{$48.8 \pm 2.0$} &-\\
&{WS-VAE} &{{$32.1 \pm 0.9$}} &{$39.9 \pm 0.9$} &{{$43.8 \pm 0.9$}}&{$49.1 \pm 2.0$} &{$65.3 \pm 1.0$}\\
&{WS-VAE-BERT} &\bm{{$41.0 \pm 0.8$}} &\bm{{$45.6 \pm 2.0$}} &\bm{{$51.2 \pm 0.8$}} &\bm{{$58.3 \pm 2.0$}} &\bm{$67.8 \pm 1.0$}\\\hline \hline
\multirow{6}{*}{Borrow} & {LSTM} &{$53.4 \pm 0.9$} &{$62.6 \pm 0.9$} &{$68.1 \pm 0.8$}&{$74.4 \pm 2.0$}  &-\\
&{SH-Net} &{$53.7 \pm 1.0$} &{$63.2 \pm 1.0$} &{$68.0 \pm 0.7$}&{$74.5 \pm 1.0$} &{$56.5 \pm 2.0$}\\
&{BERT} &{$56.7 \pm 1.0$} &{$64.1 \pm 3.0$} &{$68.5 \pm 1.0$} &{$74.6 \pm 0.4$}  &-\\\cline{2-7} 
&{S-VAE} &{$59.2 \pm 0.7$} &{$65.3 \pm 0.4$} &{$68.8 \pm 0.6$} &{$74.6 \pm 0.5$} &-\\ 
&{WS-VAE} &{{$59.5 \pm 1.0$}} &{{$66.0 \pm 0.7$}} &{{$68.9 \pm 1.0$}} &{{$74.7 \pm 0.3$}} &{$56.5 \pm 0.9$} \\ 
&{WS-VAE-BERT} &\bm{{$62.6 \pm 2.0$}} &\bm{{$68.5 \pm 1.0$}} &\bm{{$70.4 \pm 1.0$}} &{\bm{$75.9 \pm 0.7$}} &\bm{$57.5 \pm 0.8$}\\\hline 

\end{tabular}\caption{Sentence-level persuasion strategy prediction performance (Macro F1 Score) and document-level prediction performance (Accuracy). Models are trained with documents amount of 20 (81 sentences in Kiva, 99 sentences in RAOP and 59 sentences in Borrow), 50 (200 sentences in Kiva, 236 sentences in RAOP and 168 sentences in Borrow), 100 (355 sentences in Kiva, 480 sentences in RAOP and 356 sentences in Borrow), and all the training set (3512 sentences in Kiva, 1382 sentences in RAOP and 3136 sentences in Borrow). 
The results are averaged after 5 different runs, with the 95\% confidence interval.}\label{Tab:result}
\end{table*}
\textbf{Experiment Setup:}
We randomly sampled from the labeled documents to form the maximum labeled train set, the development, and test set to train and evaluate models, and we utilized all the unlabeled documents as training data as well. The data splits are shown in Table \ref{Tab:split}. We utilized NLTK \cite{Bird:2009:NLP:1717171} to split the documents into sentences and tokenize each sentence with BERT-base uncased tokenizer \cite{devlin-etal-2019-bert}. We added a special CLS token at the beginning of each sentence and a special SEP token at the end of each sentence. We used BERT \cite{devlin-etal-2019-bert} as the discriminative network, LSTM as the generative network and predictor network. The inference network is a 2-layer MLP.
We trained our model via AdamW \cite{DBLP:journals/corr/abs-1711-05101} and tuned hyper-parameters on the development set. 
\vspace{-0.05in}
\subsection{Baselines and Model Settings\footnote{Parameters details are stated in Section 5 in the Appendix.}}\vspace{-0.05in}
We compared our model on strategy classification for each sentence with several baselines:
(1) \textbf{LSTM} \cite{Hochreiter:1997:LSM:1246443.1246450}:
LSTM is utilized as the encoder for sentences. We use the last layer's hidden states as the representations of sentences to classify the persuasion strategies. Only labeled sentences are used here.
(2) \textbf{SH-Net} \cite{yang2019let}: SH-Net utilized a hierarchical LSTM to classify strategies with the supervision from both sentence-level and document-level labels, thus both labeled documents and unlabeled documents being used. We followed their implementation and modified the document-level inputs as concatenations of latent variables $y$ and $z$.  
(3) \textbf{BERT} \cite{devlin-etal-2019-bert}: We used the pre-trained BERT-base uncased model and fine-tuned it for the persuasion strategy classification. BERT only utilized labeled sentences.
(4) \textbf{S-VAE}: Sentence-level VAE applied variational autoencoders in classifications by reconstructing the input sentences while learning to classify them.
Both labeled and unlabeled sentences are used. 

\begin{figure*}[ht]
\centering
\includegraphics[width=2\columnwidth]{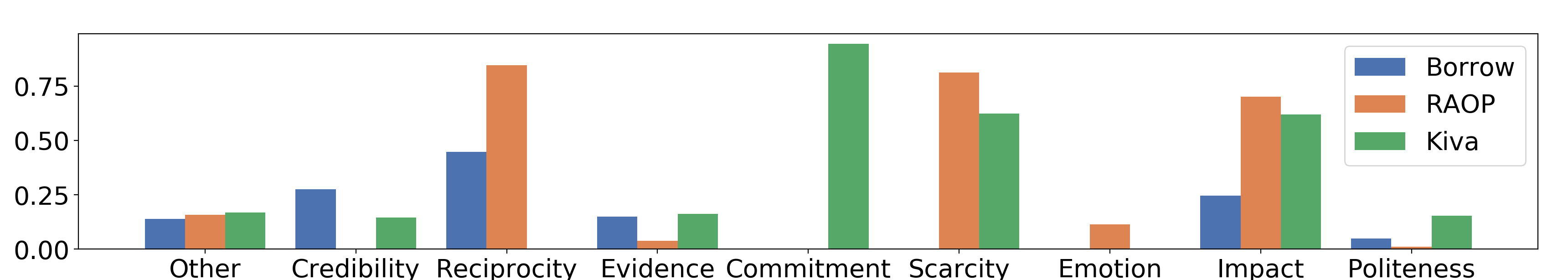}
\caption{Average attention weight learned in the predictor network for different strategies in three datasets. 
}\label{Fig:attn_dist}
\end{figure*}

\textbf{WS-VAE} denotes
our proposed weakly supervised latent variable model that made use of sentence-level labels and document-level labels at the same time, as well as reconstructing input documents. 
We further showed that our proposed WS-VAE model is orthogonal to pre-trained models like BERT as well by utilizing pre-trained BERT as the discriminative network to encode the input sentences and then using 2-layer LSTM as the generative network and predictor network, denoted as \textbf{WS-VAE-BERT},  a special case (based on pre-trained transformer models) of WS-VAE.
\vspace{-0.1in}
\subsection{Results}
\paragraph{Varying the Number of Labeled Documents} We tested the models with varying amount of labeled documents from 20 to the maximum number of labeled training documents, and summarized the results in Table \ref{Tab:result}. The simple LSTM classifier showed the worst performance over three datasets, especially when limited labeled documents were given. After simply adding document-level supervision as well as unlabeled documents, SH-Net got better Macro F1 scores as well as lower variance, showing the impact of document-level supervision on sentence-level learning. BERT fine-tuned on persuasion strategy classification tasks showed better performance than LSTM and SH-Net with limited labeled data in most cases. 

By leveraging the reconstruction of each input sentence using corresponding persuasion strategies and context latent variables, S-VAE showed a significant performance boost comparing to only utilizing indirect supervision from the document-level labels. This indicated that by incorporating the extra supervision directly from the input sentence itself, we can gain more help than hierarchical supervision from document levels. By utilizing the hierarchical latent variable model, which not only utilized the sentence reconstruction but also document-level predictions to assist the sentence-level classifications, WS-VAE outperformed S-VAE. When combining with the state-of-the-art pre-trained models like BERT, our WS-VAE-BERT achieved the best performance over three datasets. This suggests that such improvement does not only come from large pre-trained models, but also the incorporation of our hierarchical latent variable model.

Note that we also showed the document-level prediction accuracy for models that used all the labeled documents. Even though the document-level predictions were not our goals, we observed a consistent trend that higher document-level performance correlated with the higher sentence-level accuracy, suggesting that the global document-level supervision helped the sentence-level predictions.
\vspace{-0.15in}

\begin{figure}[t!]
\centering
\includegraphics[width=1\columnwidth]{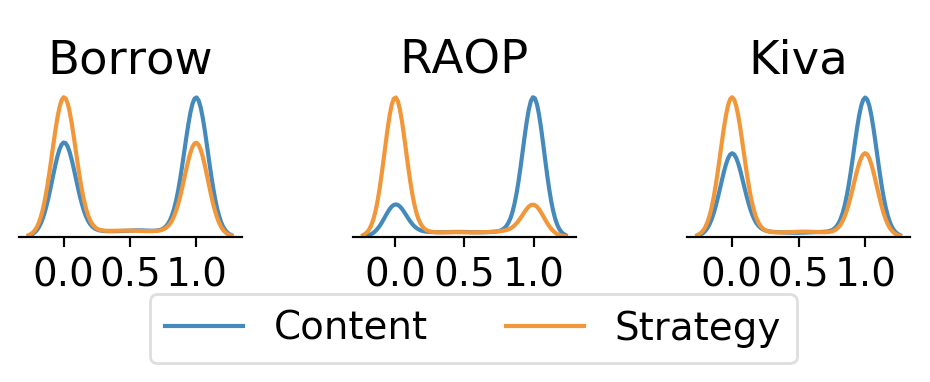}
\caption{Attention weight for content vectors and strategy vectors when predicting document-level labels in the predictor network.}\label{Fig:attn}
\end{figure}

\begin{figure}[t]
\centering
\includegraphics[width=1\columnwidth]{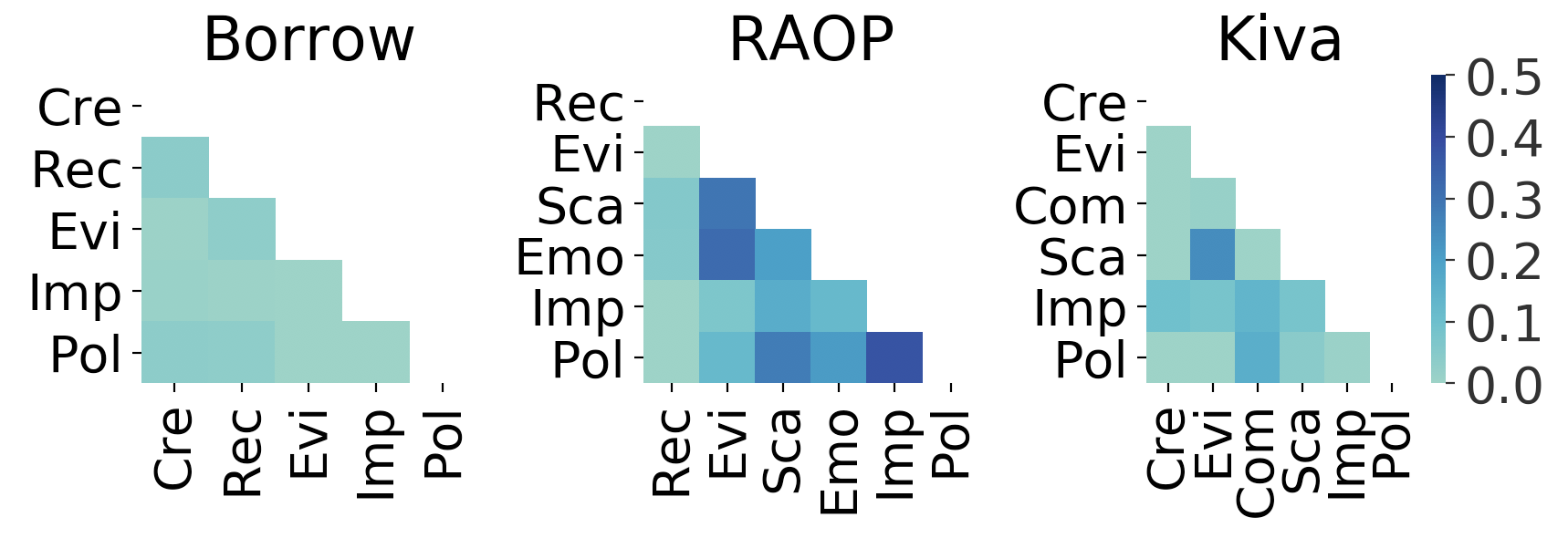}
\caption{Cosine similarities between different persuasive strategies (\textbf{Cre}dibility, \textbf{Rec}iprocity, \textbf{Evi}dence, \textbf{Com}mitment, \textbf{Sca}rcity, \textbf{Emo}tion, \textbf{Imp}act and \textbf{Pol}iteness). }\label{Fig:embeddings}
\end{figure}
\vspace{-0.1in}
\paragraph{Importance of Strategies vs Content} 
To better understand how these persuasive strategies and the text content jointly affect the success of text requests, 
we added an attention layer over content latent variable $z$ and strategy latent variable $y$ in the predictor network to visualize the importance of persuasive strategies and text content in the WS-VAE-BERT, as shown in Figure~\ref{Fig:attn}.
In all three domains, 
we found that content vectors tend to have larger weights than strategy vectors. This suggests that when people are writing requests to convince others to take action, content is relatively the more important component than persuasion strategies. However,  leveraging proper persuasive strategies can further boost the likelihood of their requests being fulfilled. 

\vspace{-0.15in}
\paragraph{Attention Weight}
We further calculated the average attention weights learned in the predictor network  (attended over strategy latent variable $y$ and content latent variable $z$ to predict the document-level labels) for different strategies in three datasets which is shown in Figure~\ref{Fig:attn_dist}. We observed that \emph{Reciprocity}, \emph{Commitment}, \emph{Scarcity} and \emph{Impact} seemed to play more important roles, while \emph{Credibility}, \emph{Evidence}, \emph{Emotion} and \emph{Politeness} had lower average attention weights, which indicated that simple superficial strategies might be more influential to overall persuasiveness in online forums than strategies that required deeper analysis.

\vspace{-0.2in}
\paragraph{Relation between Persuasive Strategies} 
To explore possible relations among different persuasive strategies, we utilized the embeddings for each persuasive strategy from the predictor network and visualized their pairwise similarities in Figure~\ref{Fig:embeddings}. All the similarities scores were below 0.5, showing those strategies in our taxonomy are generally orthogonal to each other and capture different aspects of persuasive language. However, some strategies tend to demonstrate relatively higher relations; for example, \emph{Scarcity} highly correlates with \emph{Evidence} on RAOP and Kiva, indicating that people may often use them together in their requests.

\vspace{-0.1in}
\section{Conclusion and Future Work}\vspace{-0.05in}
This work introduced a set of generic persuasive strategies based on theories on persuasion, together with a large-scale multi-domain text corpus annotated with their associated persuasion strategies. 
To further utilize both labeled and unlabeled data in real-world scenarios, 
we designed a hierarchical weakly-supervised latent variable model to utilize document-level persuasiveness supervision to guide the learning of specific sentence-level persuasive strategies. 
Experimental results showed that our proposed method outperformed existing semi-supervised baselines significantly on three datasets. 
Note that, we made an assumption that the document-level persuasiveness label only depended on the sentence-level information. However there are other factors closely related to the overall persuasiveness such as requesters/lenders' backgrounds or their prior interactions \cite{valeiras2020genre,longpre2019persuasion}. 
Future work can investigate how these audience factors further affect the predictions of both sentence- and document- level labels.
As an initial effort, our latent variable methods disentangle persuasion strategies and the content, and highlight the relations between persuasion strategies and the overall persuasiveness, which can be further leveraged by real-world applications to make textual requests more effective via different choices of persuasion strategies. 

\section{Acknowledgment}
We would like to thank Jintong Jiang, Leyuan Pan, Yuwei Wu, Zichao Yang, the anonymous reviewers, and the members of Georgia Tech SALT group for their feedback. We acknowledge the
support of NVIDIA Corporation with the donation of GPU used for this research. DY is supported in part by a grant from Google.
\bibliography{aaai21}
\bibliographystyle{aaai}

\section{Appendix}

\section{Dataset \& Annotation Details}
In different contexts, people tend to write documents with different numbers of sentences, which might be associated with different sets of persuasion strategies. 

The mean and std for number of sentences per document are 4.68 and 4.63 in Borrow, 5.10 and 4.40 in RAOP,  and 3.83 and 4.12 in Kiva. 

We recruited two graduate and two undergraduate students to label the persuasion strategies for each sentence in given documents which were randomly sampled from the whole corpus. 
Definitions and examples of different persuasion strategies were provided to the annotators.
We also conducted a training session where we asked annotators to annotate 50 example sentences and walked through them any disagreements or confusions they had.
Annotators then annotated 1200 documents by themselves independently. 

To assess the reliability of the annotated labels, the same set of documents which contained 100 documents with 400 sentences was given to annotators to label and we computed the Cohen’s Kappa coefficient. We obtained an average score of 0.538 on Kiva, 0.613 on RAOP and 0.623 on Borrow, which indicated moderate agreement and reasonable annotation quality \cite{articlecohen}.

\section{WS-VAE}
\paragraph{Sentence level VAE}
Based on prior work on semi-supervised VAEs \cite{kingma2013autoencoding}, for an input sentence $\mathbf{s}$, we assume a 
graphical model whose latent representation contains a continuous vector $\mathbf{z}$, denoting the content of a sentence, 
and a discrete persuasive strategy label $\mathbf{y}$:
\begin{align*}
    p(\mathbf{s},\mathbf{z},\mathbf{y}) = p(\mathbf{s}|\mathbf{z},\mathbf{y})p(\mathbf{z})p(\mathbf{y}).
\end{align*}

\noindent
To learn the semi-supervised VAE, we optimize the variational lower bound as our learning objective.
For unlabeled sentence, we maximize: 
\begin{equation*}
    \begin{aligned}
         \log p(\mathbf{s}) &= \log \mathbb{E}_{\mathbf{y} \sim p(\mathbf{y})}\mathbb{E}_{\mathbf{z} \sim p(\mathbf{z})}[p(\mathbf{\mathbf{s}}|\mathbf{z}, \mathbf{y})] \\
        &\geq \mathbb{E}_{\mathbf{y} \sim q(\mathbf{y}|\mathbf{s})} [\mathbb{E}_{\mathbf{z} \sim q(\mathbf{z}|\mathbf{s}, \mathbf{y})}[\log p(\mathbf{s}|\mathbf{z}, \mathbf{y})] \\
        & \quad - \text{KL}[q(\mathbf{z}|\mathbf{s},\mathbf{y})||p(\mathbf{z})]] \\
        & \quad - \text{KL}[q(\mathbf{y}|\mathbf{s}) || p(\mathbf{y})],
    \end{aligned}
\end{equation*}
where $p(\mathbf{s}|\mathbf{y},\mathbf{z})$ is a decoder (generative network) to reconstruct input sentences 
and $q(\mathbf{y}|\mathbf{s})$ is an encoder (an inference or a predictor network) to predict sentence-level labels.
For labeled sentences, the variational lower bound becomes:
\begin{equation*}
    \begin{aligned}
        \log  p(\mathbf{s}, \mathbf{y}) &= \log \mathbb{E}_{\mathbf{z} \sim p(\mathbf{z})}[p(\mathbf{s}|\mathbf{z}, \mathbf{y})p(\mathbf{y})] \\
        &\geq \mathbb{E}_{\mathbf{z} \sim q(\mathbf{z}|\mathbf{s}, \mathbf{y})} [\log p(\mathbf{s}|\mathbf{z},\mathbf{y})] \\
        &\quad - \text{KL}[q(\mathbf{z}|\mathbf{s},\mathbf{y})||p(\mathbf{z})] + \text{constant}
    \end{aligned}
\end{equation*}
In addition, for sentences with labels, we also update the inference network $q(\mathbf{y}|\mathbf{s})$ via
minimizing the cross entropy loss $\mathbb{E}_{(\mathbf{s},\mathbf{y})}[- \log q(\mathbf{y}|\mathbf{s})]$ directly.


\paragraph{Document level VAE}
Different from sentence-level VAEs, we model the input document $\mathbf{d}$ with sentences $\{\mathbf{s}^j\}_{j=1}^M=\mathbf{s}^{1:M}$ as a whole and assume that the document-level label $\mathbf{t}$ depends on the sentence-level latent variables.
Thus we obtain the document-level VAE model as:
\begin{equation*}
    \begin{aligned}
    p(\mathbf{d}, \mathbf{t}, \mathbf{y}^{1:M}, \mathbf{z}^{1:M}) = &\\ p(\mathbf{d},\mathbf{t}|\mathbf{y}^{1:M}, \mathbf{z}^{1:M})  
    &\prod_{j=1}^M p(\mathbf{y}^j) \prod_{j=1}^Mp(\mathbf{z}^j),
    \end{aligned}
\end{equation*}
where $p(\mathbf{d},\mathbf{t}|\mathbf{y}^{1:M}, \mathbf{z}^{1:M})$ is the generative model for all sentences
in the document $\mathbf{d}$ and the document label $\mathbf{t}$.
For simplicity, we further assume conditional independence between the sentences $\mathbf{s}^{1:M}$ in  $\mathbf{d}$ and its label $\mathbf{t}$ given the latent variables:
\begin{align*}
    p(\mathbf{d},\mathbf{t}|\mathbf{y}^{1:M}, \mathbf{z}^{1:M}) = &\\ p(\mathbf{t}|\mathbf{y}^{1:M}, \mathbf{z}^{1:M}) \nonumber
    & \prod_{j=1}^M p(\mathbf{s}^j|\mathbf{y}^j, \mathbf{z}^j).
\end{align*}
\noindent
Since the possible number of the sentence label combinations is huge, simply computing the marginal probability becomes intractable. 
Thus we optimize the evidence lower bound. 
By using mean field approximation \cite{DBLP:journals/corr/abs-1802-06126}, we factorize the posterior distribution as:
\vspace{-0.2cm}
\begin{equation*}
\begin{aligned}
   & q(\mathbf{z}^{1:M}, \mathbf{y}^{1:M} | \mathbf{d}, \mathbf{t}) \\
   & \quad \quad = q(\mathbf{z}^{1:M}|\mathbf{y}^{1:M},\mathbf{s}^{1:M}, \mathbf{t}) q(\mathbf{y}^{1:M}|\mathbf{s}^{1:M}, \mathbf{t}) \\
   & \quad \quad = \prod_{j=1}^M q(\mathbf{z}^j|\mathbf{y}^j,\mathbf{s}^j, \mathbf{t}) \prod_{j=1}^M q(\mathbf{y}^j|\mathbf{s}^j, \mathbf{t}),
\end{aligned}
\end{equation*}
\noindent
\noindent
That is, the posterior distribution of latent variables $\mathbf{y}^j$ and $\mathbf{z}^j$ only depends on
the sentence $\mathbf{s}^j$ and the document label $\mathbf{t}$.
For documents without sentence labels, the variational lower bound $U(\mathbf{d}, \mathbf{t})$ is:
\begin{align*}
    & \log p(\mathbf{d}, \mathbf{t}) 
    =  \log \mathbb{E}_{\mathbf{y} \sim p(\mathbf{y})}\mathbb{E}_{\mathbf{z}\sim p(\mathbf{z})} [p(\mathbf{t}|\mathbf{z}^{1:M}, \mathbf{y}^{1:M}) \nonumber\\
    & \quad\quad\quad\quad\quad\quad \prod_{j=1}^M p(\mathbf{s}^j|\mathbf{z}^j, \mathbf{y}^j)\prod_{j=1}^M p(\mathbf{y}^j) \prod_{j=1}^M p(\mathbf{z}^j)] \nonumber\\
    &\geq \mathbb{E}_{\mathbf{y}^{1:M} \sim q(\mathbf{y}^{1:M}|\mathbf{s}^{1:M}, \mathbf{t})} [\mathbb{E}_{\mathbf{z}^{1:M} \sim q(\mathbf{z}^{1:M}|\mathbf{s}^{1:M}, \mathbf{y}^{1:M}, \mathbf{t})} \nonumber\\
    &\quad [\log p(\mathbf{t}|\mathbf{y}^{1:M}, \mathbf{z}^{1:M}) + \sum_{i=1}^N \log p(\mathbf{s}^j|\mathbf{y}^j, \mathbf{z}^j)] \nonumber\\
    &\quad\quad\quad\quad - \sum_{j=1}^M \text{KL}[q(\mathbf{z}^j|\mathbf{s}^j, \mathbf{y}^j, \mathbf{t})||p(\mathbf{z}^j)]] \nonumber\\
    &\quad\quad\quad\quad - \sum_{j=1}^M \text{KL}[q(\mathbf{y}^j|\mathbf{s}^j, \mathbf{t})||p(\mathbf{y}^j)]  \nonumber\\
    &\quad\quad\quad\quad = U(\mathbf{d}, \mathbf{t})
\end{align*}
For document with sentence labels, the variational lower bound can be adapted from above as:
\begin{align*}
    & \log p(\mathbf{d}, \mathbf{t}, \mathbf{y}^{1:M}) \nonumber\\
    &=  \log \mathbb{E}_{\mathbf{z}\sim p(\mathbf{z})} [p(\mathbf{t}|\mathbf{z}^{1:M}, \mathbf{y}^{1:M}) \nonumber\\
    & \quad\quad\quad\quad\quad\quad \prod_{j=1}^M p(\mathbf{s}^j|\mathbf{z}^j, \mathbf{y}^j)\prod_{j=1}^M p(\mathbf{y}^j) \prod_{j=1}^M p(\mathbf{z}^j)] \nonumber\\
    &\geq \mathbb{E}_{\mathbf{z}^{1:M} \sim q(\mathbf{z}^{1:M}|\mathbf{s}^{1:M}, \mathbf{y}^{1:M}, \mathbf{t})} \nonumber\\
    &\quad [\log p(\mathbf{t}|\mathbf{y}^{1:M}, \mathbf{z}^{1:M}) + \sum_{i=1}^N \log p(\mathbf{s}^j|\mathbf{y}^j, \mathbf{z}^j)] \nonumber\\
    &\quad -\sum_{j=1}^M \text{KL}[q(\mathbf{z}^j|\mathbf{s}^j, \mathbf{y}^j, \mathbf{t})||p(\mathbf{z}^j)] + \text{constant} \nonumber\\
    &\quad= L(\mathbf{d}, \mathbf{t}, \mathbf{y}^{1:M}) + \text{constant}
\end{align*}
Combining the loss for document with and without sentence labels,
we obtain the overall loss function:
\begin{align*}
    L =&  \quad \mathbb{E}_{\mathbf{d} \in \mathbf{D}_U} U(\mathbf{d}, \mathbf{t}) +  \mathbb{E}_{\mathbf{d} \in \mathbf{D}_L} L(\mathbf{d}, \mathbf{t}, \mathbf{y}^{1:M}) \nonumber\\
    & \quad + \alpha \cdot \mathbb{E}_{\mathbf{d} \in \mathbf{D}_L} \prod_{j=1}^{M} \log q(\mathbf{y}^j| \mathbf{s}^j, \mathbf{t})
\end{align*}
Here,  $\mathbb{E}_{\mathbf{d} \in \mathbf{D}_L} \prod_{j=1}^{M} \log q(\mathbf{y}^j| \mathbf{s}^j, \mathbf{t})$
represents the discriminative loss for sentences with persuasive strategy labels and $\alpha$ controls the trade-off between generative loss and discriminative
loss.

\section{Threshold on KL Divergence} \label{Sec:Y_threshold}
\citet{yang2017improved} found that VAEs might easily get stuck in two local optimums: the KL term on $\mathbf{y}$ is very large and all samples collapse to one class or the KL term on $\mathbf{y}$ is very small and $q(\mathbf{y}|\mathbf{s})$ is close to the prior distribution. Thus we minimize the KL term only when it is larger than a threshold $w$: 
\begin{equation*}
    \text{KL}_\mathbf{y} = \mathbf{max}(w, \text{KL}[q(\mathbf{y}|\mathbf{s})||p(\mathbf{y})])
\end{equation*}

\section{Influence of the Trade-off Weight $\alpha$}
The overall loss function of our proposed weakly-supervised hierarchical latent variable model is:

\begin{align*}
    L =&  \quad \mathbb{E}_{\mathbf{d} \in \mathbf{D}_U} U(\mathbf{d}, \mathbf{t}) +  \mathbb{E}_{\mathbf{d} \in \mathbf{D}_L} L(\mathbf{d}, \mathbf{t}, \mathbf{y}^{1:M}) \nonumber\\
    & \quad + \alpha \cdot \mathbb{E}_{\mathbf{d} \in \mathbf{D}_L} \prod_{j=1}^{M} \log q(\mathbf{y}^j| \mathbf{s}^j, \mathbf{t})
\end{align*}

Here, the $\alpha$ is a parameter that controls the balance of reconstruction loss and supervised sentence classification loss. When $\alpha$ is small, the sentence level classifications are not well learned. When $\alpha$ is large, the model tends to only learn the sentence level classification tasks and ignore the reconstructions and document level predictions.  
In experiments, we set $\alpha$ to 5 through a grid search from the set $\{1, 5, 10, 20\}$.

\section{Model Implementation Details}

\subsection{S-VAE}
For \textbf{S-VAE} - 
the sentence-level latent variable model, which applies variational autoencoderes in sentence-level classifications by reconstructing the input sentences while learning to classify them, which encourages the model to assign input sentences to a label $y$ such that the reconstruction loss is low. S-VAE is a special case (only performing operations at sentence levels) of our proposed WS-VAE. The weight for the reconstruction term is 1, the weight for the classification term is 5 and the weight for KL divergence terms are annealing from a small value to 1 through the training process. The learning rate is 0.001. 

\subsection{WS-VAE}
For \textbf{WS-VAE} -
our proposed weakly supervised latent variable model, takes advantage of sentence-level labels and document-level labels at the same time, as well as reconstructing input documents. 
The weight for the reconstruction term is 1, the weight for the classification term is 5, the weight for KL divergence terms are annealing from a small value to 1 through the training process, and the weight for predictor term is 0.5. The threshold for KL regularization on $q(y|s)$ is 1.2. The learning rate is 0.001. 

\subsection{WS-VAE-BERT}
For \textbf{WS-VAE-BERT} -  a special case (based on pre-trained transformer models) of WS-VAE, combines ES-VAE with recent pre-trained BERT. The weight for the reconstruction term is 1, the weight for the classification term is 5, the weight for KL divergence terms are annealing from a small value to 1 through the training process, and the weight for predictor term is 0.1. The threshold for KL regularization on $q(y|s)$ is 1.2.  The learning rate is 0.00001.

\begin{table}[t]
\centering
\begin{tabular}{c|c|c}
\hline
\textbf{Datasets}       & \textbf{Threshold on y} & \textbf{Macro F1} \\ \hline \hline
\multirow{3}{*}{Kiva}   & 0                       & 0.228             \\ \cline{2-3}
                        & 1.2                     & 0.315             \\ \cline{2-3}
                        & 2.0                     & 0.305             \\ \hline
\multirow{3}{*}{RAOP}   & 0                       & 0.274             \\ \cline{2-3}
                        & 1.2                     & 0.321             \\ \cline{2-3}
                        & 2.0                     & 0.316             \\ \hline
\multirow{3}{*}{Borrow} & 0                       & 0.485             \\ \cline{2-3}
                        & 1.2                     & 0.595             \\ \cline{2-3}
                        & 2.0                     & 0.542  \\ \hline           
\end{tabular}
\caption{Macro F1 Score with different threshold on
y in KL regularization term for SH-VAE. Models are trained on three datasets with 20 labeled documents (81 sentences in kiva, 99 sentences in RAOP and 59 sentences in Borrow).} \label{Tab:Vary_K}
\end{table}

\section{Impact of Variational Regularization}
To show the importance of variational regularization on the latent variable $y$ (the threshold on KL divergence $w$) mentioned in Section~\ref{Sec:Y_threshold}, we performed ablation study for the KL term for $y$. We tested WS-VAE with different values of threshold on three datasets using 20 labeled documents and the results were shown in Table~\ref{Tab:Vary_K}. When the threshold is small like 0, which meant we added large regularization on y, the performance  is bad because the  $q(y|s)$ was so close to estimated prior distributions and barely learned from objective functions. When the threshold was large like 2, which meant there did not exist any regularization on $y$, we got lower F1 scores as well. When there is a appropriate threshold such as 1.2 to offer regularization, WS-VAE could achieve the best performance.

\begin{figure}[t!]
\centering
\includegraphics[width=0.8\columnwidth]{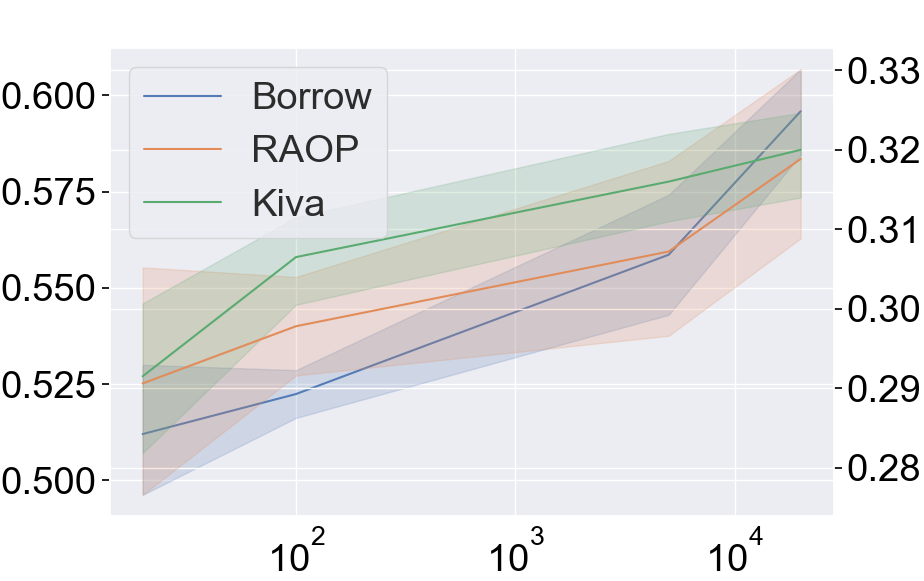}
\caption{\small{Macro F1 scores with 20 documents with sentence labels and different numbers of documents without sentence labels for WS-VAE. Results on Borrow follow the left y-axis, while RAOP and Kiva follow the right y-axis.}}\label{Fig:unlabeled}
\end{figure}

\section{Varying the Number of Unlabeled Documents}
We visualized WS-VAE's performances on three datasets when varying the amount of unlabeled data in Figure \ref{Fig:unlabeled}: macro F1 scores increased with more unlabeled data, demonstrating the effectiveness of the introduction of unlabeled sentences, and our hierarchical weakly-supervised model. 

\end{document}